\documentclass[preprint,12pt]{elsarticle}

\usepackage{bbm}
\usepackage{subcaption}
\usepackage[english]{babel}
\usepackage[utf8]{inputenc} 
\usepackage[T1]{fontenc}    
\usepackage{hyperref}       
\usepackage{url}            
\usepackage{booktabs}       
\usepackage{amsfonts}       
\usepackage{nicefrac}       
\usepackage{microtype}      
\usepackage{algorithm}
\usepackage{algorithmic}
\usepackage{makecell}
\usepackage{colortbl} 
\definecolor{mygray}{gray}{.9}
\usepackage{graphicx}
\usepackage{subcaption}
\usepackage{amsmath}
\usepackage{amsthm}
\usepackage{color}
\usepackage{multirow}
\usepackage{mathrsfs}
\usepackage{float}
\usepackage[misc]{ifsym}
\usepackage{caption}
\usepackage{amsthm}
\usepackage{graphicx}
\usepackage{pythonhighlight}
\usepackage{bm}



\journal{Elsevier}

\begin{document}

\begin{frontmatter}



\title{3D Hand Mesh-Guided AI-Generated Malformed Hand Refinement with Hand Pose Transformation via Diffusion Model}

\author{\Letter Chen-Bin Feng\fnref{label1}}\ead{fengchenbinjacob@gmail.com}
\address[label1]{Department of Computer and Information Science, University of Macau, Macau, China \fnref{label1}}
\address[label2]{ShanghaiTech University, Shanghai, China \fnref{label2}}
\author{Kangdao Liu\fnref{label1}}
\author{Jian Sun\fnref{label1}}
\author{Jiping Jin\fnref{label2}}
\author{Yiguo Jiang\fnref{label1}}
\author{\Letter Chi-Man Vong\fnref{label1}}


\begin{abstract}
The malformed hands in the AI-generated images seriously affect the authenticity of the images. To refine malformed hands, existing depth-based approaches use a hand depth estimator to guide the refinement of malformed hands. Due to the performance limitations of the hand depth estimator, many hand details cannot be represented, resulting in errors in the generated hands, such as confusing the palm and the back of the hand. To solve this problem, we propose a 3D mesh-guided refinement framework using a diffusion pipeline. We use a state-of-the-art 3D hand mesh estimator, which provides more details of the hands. For training, we collect and reannotate a dataset consisting of RGB images and 3D hand mesh. Then we design a diffusion inpainting model to generate refined outputs guided by 3D hand meshes. For inference, we propose a double check algorithm to facilitate the 3D hand mesh estimator to obtain robust hand mesh guidance to obtain our refined results. Beyond malformed hand refinement, we propose a novel hand pose transformation method. It increases the flexibility and diversity of the malformed hand refinement task. We made the restored images mimic the hand poses of the reference images. The pose transformation requires no additional training. Extensive experimental results demonstrate the superior performance of our proposed method.
\end{abstract}

\end{frontmatter}











\section{Introduction}\label{sec:introduction}
\begin{figure}
    \centering
    \includegraphics[width=12cm]{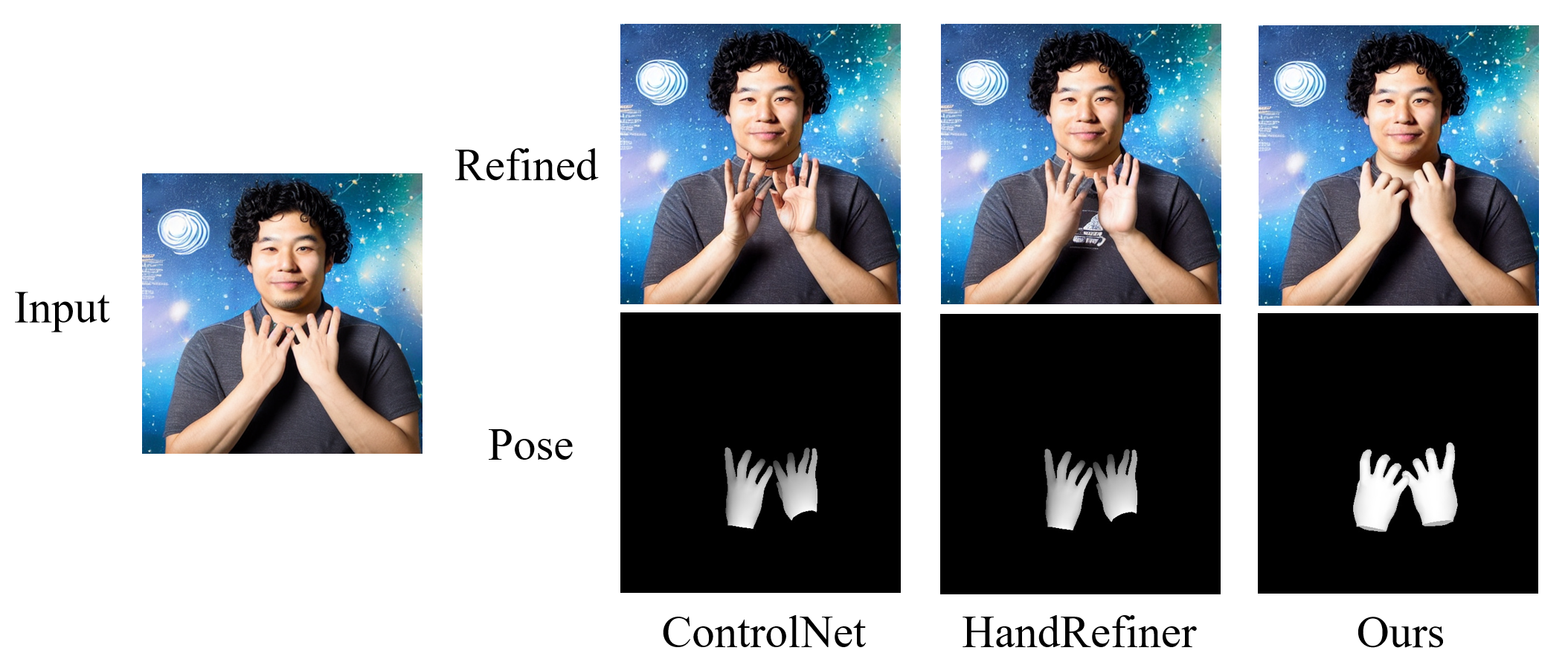}
    \caption{\textbf{The visual comparison of input malformed hand image by Stable Diffusion \cite{rombach2022high}, refined result by ControlNet \cite{zhang2023adding}, refined result by HandRefiner \cite{lu2024handrefiner}, and refined result by our method.} The former two methods use depth pose, which lacks details and is thus hard to guide the refining model to generate the palm or the back of the hand. Our method uses 3D hand mesh that provides rich details for generating realistic hands.}
    \label{fig:comp1}
    \vspace{10mm}
\end{figure}

Stable diffusion \cite{rombach2022high} have recently attracted a lot of attention. However, the generation results often have large artifacts when it comes to images with human hands. The generated deformed hands cause the results to be unrealistic. In this situation, HandRefiner \cite{lu2024handrefiner} makes use of stable diffusion inpainting and hand pose estimation \cite{lin2021mesh} to refine the malformed hand. It uses generated hand depth maps as guidance to refine malformed hands. However, this approach performs poorly when the hand pose estimator \cite{zhang2020mediapipe} does not provide enough detail about the structure of the hand. In particular, The depth map-based hand pose estimator fails to provide details that tell the model whether the hand is the palm or the back of the hand. This motivates us to use a better hand pose estimation approach and more real-world hand data to solve such problems. We choose the 3D hand mesh as a substitution for hand depth, which shows better hand pose representation performance. To implement our approach, we design training and inference pipelines and organize training datasets for model training. Specifically, we develop a conditional diffusion inpainting pipeline. We input the estimated 3D hand mesh as conditions into a diffusion model \cite{song2020denoising} that generates refined images with clear hands. We mask the hand region and use the background region as condition to estimate residual noise for diffusion model. During inference, we first estimate 3D hand meshes and then generate refined images conditioned on the meshes. A visual comparison are shown in Figure \ref{fig:comp1}. It shows that our method generates more clear hand guidance and more realistic hands.

\begin{figure}
    \centering
    \includegraphics[width=12cm]{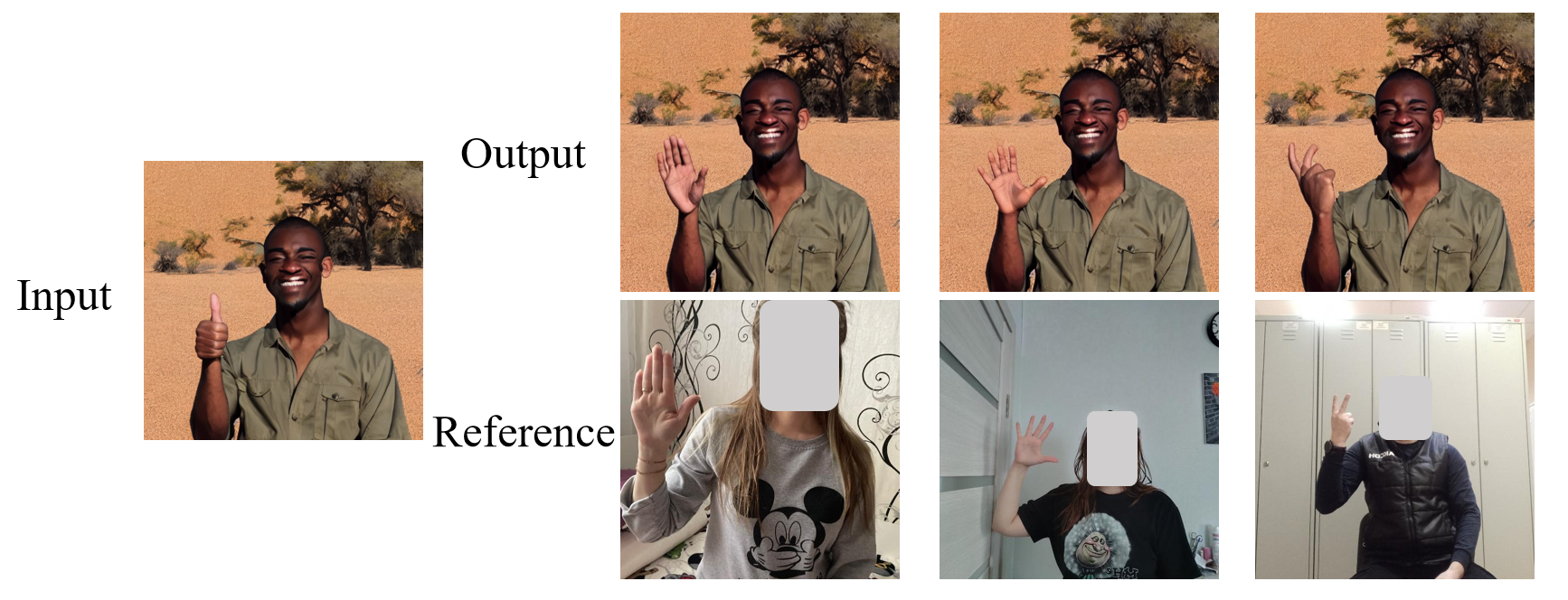}
    \caption{\textbf{The visual result of our proposed hand pose transformation.} The outputs change pose according to reference images. Faces of the real person are obscured due to privacy protection.}
    \label{fig:posetrans1}
    \vspace{10mm}
\end{figure}

In addition, the existing approach refines human hands based on the estimated hand pose of the input images, which is not flexible. It cannot generate arbitrary hand poses followed by user preferences. This motivates us to propose a method for hand pose transformation. We generate arbitrary hand pose image by aligning another hand pose from the reference image and conduct hand image generation. Using our method, the user can change an arbitrary hand pose of a refined image by providing a reference hand pose image. This transform method does not require additional training. A visual result are shown in Figure \ref{fig:posetrans1}. We can see that our method can render different poses according to references. The major contributions of this paper can be concluded as follows:

\begin{itemize}
    \item We propose a diffusion-based 3D hand mesh-guided malformed hand refinement framework. 
    \item We reannotate, filter, and organize a hand refinement dataset. It contains 24,411 3D hand meshes and hand images. 
    \item We propose a double check algorithm that increases the confidence of hand mesh estimation and benefits the robustness of hand refinement. 
    \item We propose a novel hand pose transformation algorithm without additional training. To the best of our knowledge, this is the first hand pose transformation method for malformed hand refinement.

\end{itemize}

\section{Related works}
\subsection{Hand Image Generation}
Image generation \cite{rombach2022high,zhou2024adaptive,song2025attridiffuser,zhen2025token} is an important computer vision task. Most of the image generation methods can generate human images, however, some of the generated images have poor human hands. Methods for human hand image generation aim to improve the quality of generated hand images. Handcraft \cite{qin2024handcraft}, uses two kinds of human pose templates, the open palm and the fist, to fix the badly shaped hand. Hand1000 \cite{zhang2024hand1000} uses 1000 hand images to fine-tune the stable diffusion model, resulting in better hand image generation capability. It cannot generate arbitrary hand pose images. AttentionHand \cite{park2024attentionhand} uses hand images with hand mesh to train a text-to-image model similar to Controlnet \cite{zhang2023adding}. It generates images with hand poses given a hand mesh, but it cannot refine malformed hand images. GraspDiffusion \cite{kwon2024graspdiffusion} and ManiVideo \cite{pang2024manivideo} both synthesize the interaction between the hand and the object of the whole body. VTON \cite{liang2024vton} provides a hand-aware virtual try-on solution. RHands \cite{wang2025rhands}  provides a style-guided hand refinement pipeline. GivingHand \cite{pelykh2024giving} uses human pose and segmentation map to generate text-to-image. MGHanD \cite{eum2025mghand} uses LoRA weights and discriminator weights to guide hand image generation.  RealisHuman \cite{wang2025realishuman} not only refines hands but also refines faces. HandDrawer \cite{fu2025handrawer} uses the whole-body depth map as guidance for hand image generation. Among them, our method fills the gap of malformed hand refinement and arbitrary hand gesture transformation.

\subsection{2D and 3D Hand Dataset}
The 2D hand dataset includes but is not limited to hand keypoint pose, hand segmentation, and hand depth. While the 3D hand datasets usually use 3D hand meshes \cite{gao2024progressively} and can be rendered to 2D mesh maps given a camera pose. Mediapipe \cite{zhang2020mediapipe} is a general perceptual pipeline that can be used to detect human pose and hand pose. MANO \cite{romero2022embodied} is a mainstream 3D hand mesh rendering method. Mesh Graphomer \cite{lin2021mesh} is a hand mesh estimation approach that can output hand depth maps. However, it often fails to provide hand details due to the limited training data. FreiHAND \cite{zimmermann2019freihand} uses a multi-camera system to give high-quality and accurate annotations for hand images. RHD \cite{zimmermann2017learning} provides a synthesis human hand dataset with depth annotations. HaGrid \cite{kapitanov2024hagrid} is a real-world human hand dataset with bounding boxes of hands and gesture classes. StaticGesture \cite{synthesisai2023static} is a synthesis human hand dataset with segmentation, depth, and normal map annotations. HIC \cite{tzionas2016capturing} is a hand and object dataset with mesh, Kinematic skeleton, and degrees of freedom (DoF) annotations. Interhand \cite{moon2020interhand2} provides a large-scale real-world hand dataset with 3D mesh annotations. ReIH \cite{moon2023dataset} is a relighted hand dataset in different backgrounds. InterWild \cite{moon2023bringing} is an 3D hand pose estimation method.  Hand dataset with 3D hand mesh annotation provides training data for our hand generation task.

\section{Method}
\subsection{Framework}
For our 3D mesh-guided malformed hand refinement task, we input an image containing a malformed hand into the model to generate a refined output. Internally, the model first uses a 3D mesh estimator to reconstruct the 3D geometry of the hand. Subsequently, the estimated 3D hand mesh is padded to create a bounding box mask that mark the hand region. Finally, the original input image, the estimated 3D hand mesh, and the generated mask are jointly fed into the refinement module of the model, which then produces the final refined output image. Note that our 3D meshes are rendered to 2D by the default camera pose.

For training, we use hands images, mesh maps and masks to get refined output. In our dataset, the input images have ground truth meshes. We train a diffusion model to enable it to learn generating realistic human hands conditioned on given hand poses. During the training process, we only need to use clear normal hands and poses to teach the model. During inference, we can use malformed hands and normal hand poses predicted by the pose estimator that are closest to the malformed hands to enable the model to generate hands. After training, the model learns to predict the refined output image given images, pose meshes, and masks. 

\subsection{Training Dataset Preparation}
\begin{figure*}
	\centering
	\includegraphics[width=12cm]{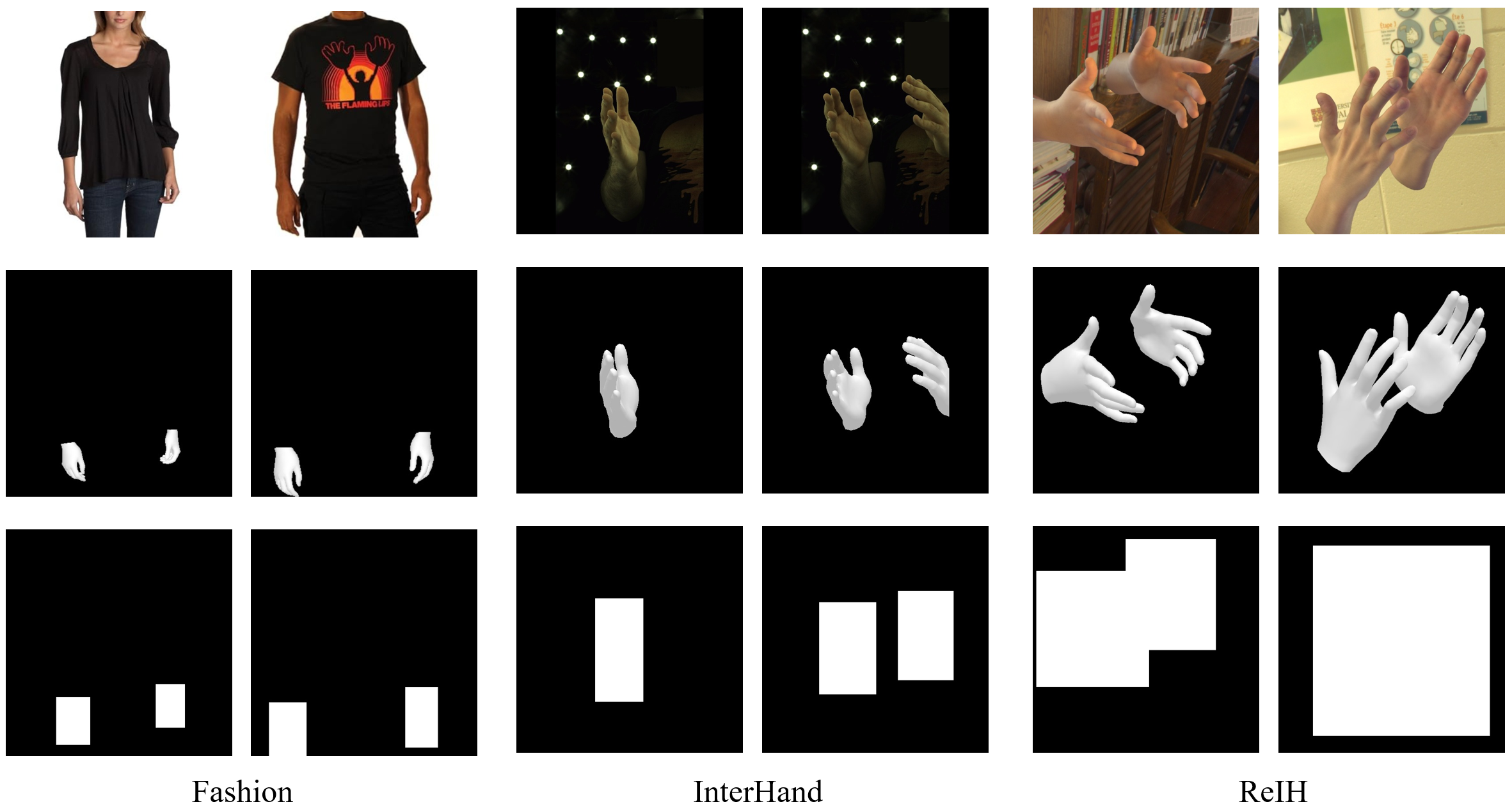}
	\caption{\textbf{Training dataset visualization.} We have three sub-datasets from different scenes.}
	\label{fig:trainv}
	\vspace{10mm}
\end{figure*}

Previous works \cite{lu2024handrefiner,wang2025rhands} utilize hand depth as guidance for malformed hand refinement. They are limited by the coarseness of depth maps and cannot provide sufficiently clear details. Therefore, we train our model using 3D hand mesh as hand pose guidance. Specifically, we sample images and ground-truth hand meshes from two 3D hand mesh datasets: InterHand \cite{moon2020interhand2} and ReIH \cite{moon2023dataset}. Our sampling strategy is to select valid images suitable for human hand image refinement while filtering out low-quality samples, such as those featuring hands occluded by objects. The InterHand dataset contains human hand images in a lab environment. It lacks background variety. ReIH, conversely, contains hand images in various backgrounds with relighted appearances. To enhance data diversity, we annotate the Fashion sub-dataset sampled from FashionIQ \cite{wu2021fashion}. We first filter out low-quality images and then annotate hand meshes using a state-of-the-art estimation approach InterWild \cite{moon2023bringing}. Finally, we manually exclude erroneous predictions. In total, the Fashion sub-dataset contains 5,021 triplets (image, mesh, mask), the InterHand sub-dataset contains 5,472 triplets, and the ReIH sub-dataset contains 13,918 triplets. A visualization of our training dataset is shown in Figure \ref{fig:trainv}.

\begin{figure*}
\centering
  \includegraphics[width=14cm]{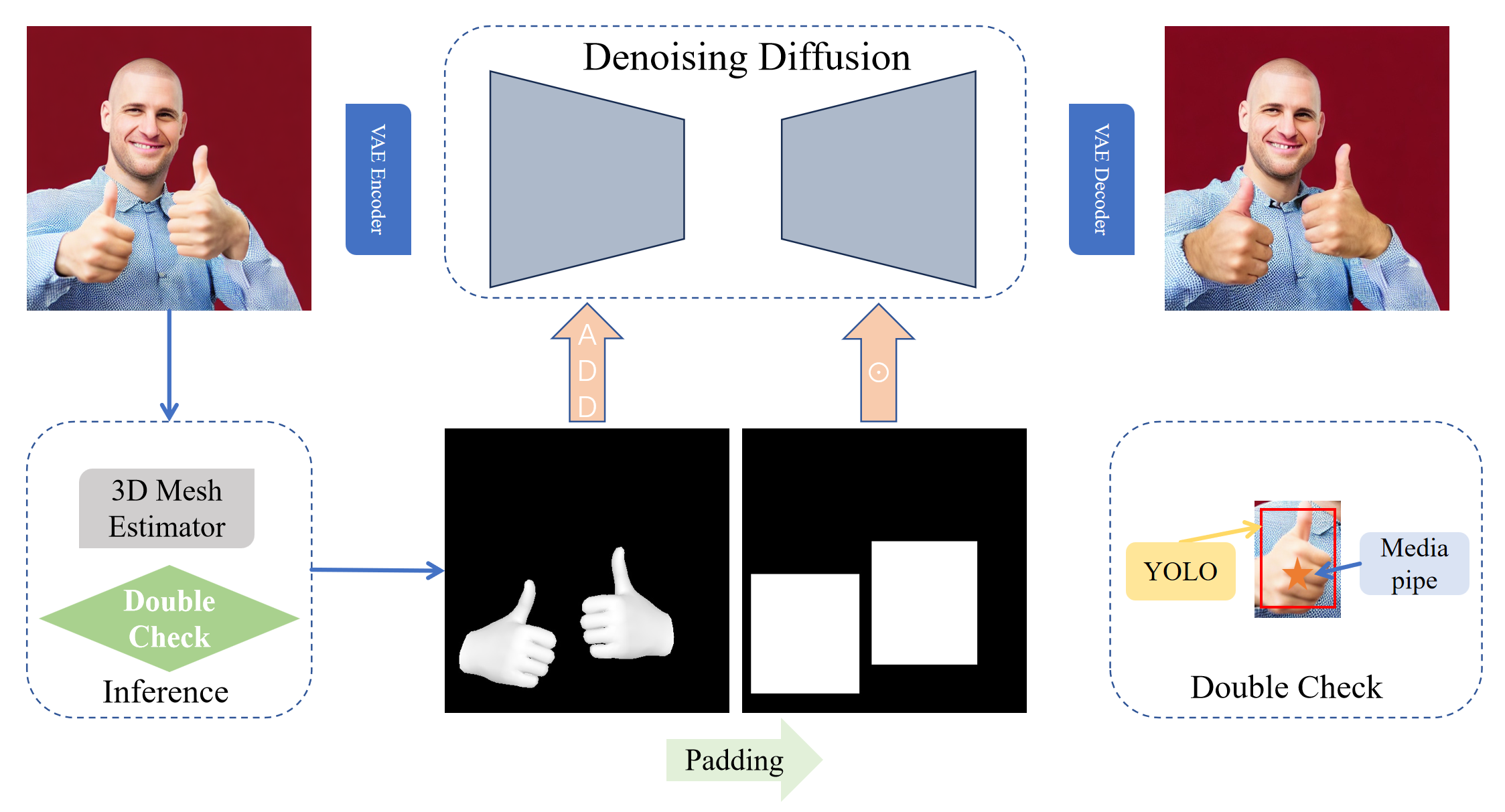}
  \caption{\textbf{The illustration of our 3D hand mesh guided hand refinement pipelines combined with double check algorithm.} The 3D mesh estimator are used during inference.}
  \label{fig:teaser} 
    \vspace{10mm}
\end{figure*}

\subsection{3D Hand Mesh Guided Diffusion Inpainting}
\textbf{3D Hand Mesh Guidance.} To ensure the refined hand is both precise and visually distinct, we employ 3D hand meshes to guide the diffusion inpainting model. Unlike depth-based guidance, these 3D meshes offer richer geometric details, enabling more accurate hand reconstruction. We use a state-of-the-art 3D hand pose estimation method, Interwild \cite{moon2023bringing}, to predict hand meshes. It provides detailed 3D hand structure. We use rendered 2D grayscale map of the 3D hand mesh for our guidance. The size of the corresponding map is of $H\times W$. $H$ and $W$ stands for height and width of the input image, respectively. Then, we pad the mesh map to get the box-shaped mask, which is used to provide the model with the inpainting location. Our pipeline is illustrated in Figure \ref{fig:teaser}.

\textbf{Diffusion Inpainting.} We input images with malformed hands, generated meshes, masks, along with text prompts into the Stable Diffusion inpainting model. The mesh guidance is fed into an efficient adapter network similar to ControlNet \cite{zhang2023adding}. For the diffusion inpainting process, we follow the design of Repaint \cite{lugmayr2022repaint}. Specifically, we utilize the DDIM sampler \cite{song2020denoising} for refining hands and use noisy backgrounds as the unmasked region. In the inpainting process, we first transform the input image to $x_0^{\text{known}}$ by adding noise in reverse step $i \in \{T, T-1, …, 2\}$. The generated result is as follows,
\begin{equation}
    x_{\tau_{i-1}}^{\text{known}} \sim N(\sqrt{\alpha_{\tau_{i-1}}}x_0^{\text{known}}, (1-\alpha_{\tau_{i-1}})\textbf{I}),
    \label{eq:noisemodel}
\end{equation}
where $N(\cdot)$ stands for the Gaussian distribution. $\alpha_{\tau_{i-1}}$ stands for a variance schedule function \cite{song2020denoising}, $\textbf{I}$ stands for identity matrix.

Then, the U-Net denoise model is used to estimate the noise. We use the DDIM sampler to get the hand region and use $x_{\tau_{i-1}}^{\text{known}}$ for unmasked region. We obtain $x_{\tau_{i-1}}$ as follows,
\begin{equation}
    x_{\tau_{i-1}} = m\odot \text{DDIM}(\epsilon_\theta(x_{\tau_i}, x_{\text{mask}}, Mesh)) + (1-m)\odot {x_{\tau_{i-1}}^{\text{known}}},
    \label{eq:ddim}
\end{equation}

where $m$ is the mask in the hand refinement stage. $x_{\text{mask}}$ denotes the masked image. They are combined with the noise vector $x_{\tau_i}$ as the inputs to the Stable Diffusion model. $Mesh$ is the corresponding 3D hand meshes. $\text{DDIM}(\cdot)$ stand for a DDIM sampling step, while $\epsilon_\theta(\cdot)$ is the U-Net denoise model. $\odot$ denotes the Hadamard product. The last DDIM step is processed without masks as follows, 
\begin{equation}
    x_{\tau_{i-1}} = \text{DDIM}(\epsilon_\theta(x_{\tau_i}, Mesh)),
    \label{eq:ddim}
\end{equation}
It is used to make the hand region and the background region harmonious. Finally, the generated features are fed into the VAE decoder to get the images with refined hands.

\begin{figure*}
\centering
  \includegraphics[width=11cm]{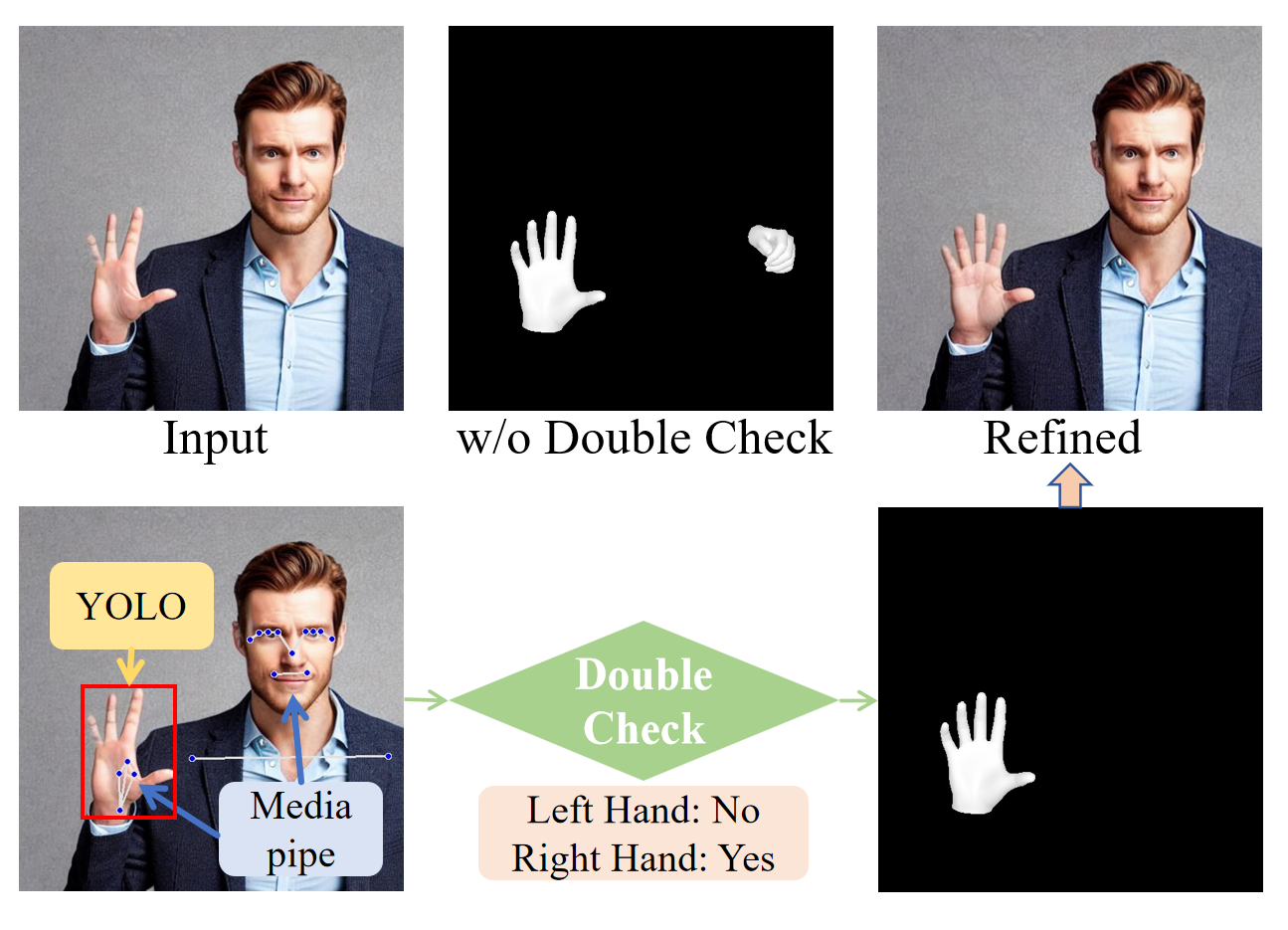}
  \caption{\textbf{The illustration of the double check.} It provides more credible results for the hand image generation model.}
  \label{fig:double} 
    \vspace{10mm}
\end{figure*}

\subsection{Double Check}
In our experiment, we notice that the hand mesh estimator proposed in \cite{moon2023bringing} has a drawback: it sometimes predicts two hands, with one of the predictions being incorrect when applied to images containing a single hand. Therefore, we propose an empirical algorithm called double check to prevent this situation from occurring, as shown in Figure \ref{fig:double}. Specifically, we use two expert hand prediction models to determine whether left hand or right hand exists in the image. The two models are YOLO \cite{redmon2016you} and Mediapipe \cite{zhang2020mediapipe}. While YOLO can predict hand bounding boxes, it lacks the capability to classify whether a hand is left or right. In contrast, Mediapipe can distinguish between left and right hands but cannot determine the presence or absence of a hand. Combining them, we get a left-right-hand existence judging algorithm. The judging criterion we use is that when the coordinates of the left hand keypoints or the right hand keypoint predicted by Mediapipe are within the range of the YOLO bounding box, we consider that the left or right hand exists. We propose the general algorithm in Algorithm \ref{alg:double}. Note that in the algorithm, 1,2,3,4 represent keypoints of the left hand while 5,6,7,8 represent keypoints of the right hand. In the box mask $M$, the box regions are given the value 1 while other regions are 0.

\begin{algorithm}[tb]
    \caption{Double Check Algorithm}
    \label{alg:double}
    \textbf{Input}: Pretrained hand detector models YOLO and Mediapipe, hand mesh predictor Interwild, and input malformed hand image $img$.\\
    \textbf{Output}: Predicted hand mesh.
    \begin{algorithmic}[1] 
        \STATE Box Mask M  = YOLO ($img$). ($\text{M} \in \{0, 1\}$)
        \STATE Key points coordinates $x_i, y_i$  = Mediapipe ($img$). ($i \in \{1, 2, 3, 4, 5, 6, 7, 8\}$)
        \STATE \text{LEFT\_EXIST}, \text{RIGHT\_EXIST} $\gets$ \text{TRUE}, \text{TRUE} 

        \FOR{$i \in \{1, 2, 3, 4\}$}

            \IF{M($x_i, y_i$) $\neq$ 1}
                \STATE \text{LEFT\_EXIST} $\gets$ \text{FALSE}
            \ENDIF
        \ENDFOR
        \FOR{$i \in \{5, 6, 7, 8\}$}
            \IF{M($x_i, y_i$) $\neq$ 1}
                \STATE \text{RIGHT\_EXIST} $\gets$ \text{FALSE}
            \ENDIF
        \ENDFOR
        \STATE Hand Mesh $\gets$ Interwild ($img$, \text{LEFT\_EXIST}, \text{RIGHT\_EXIST})
    \end{algorithmic}
\end{algorithm}

\begin{figure*}
\centering
  \includegraphics[width=12cm]{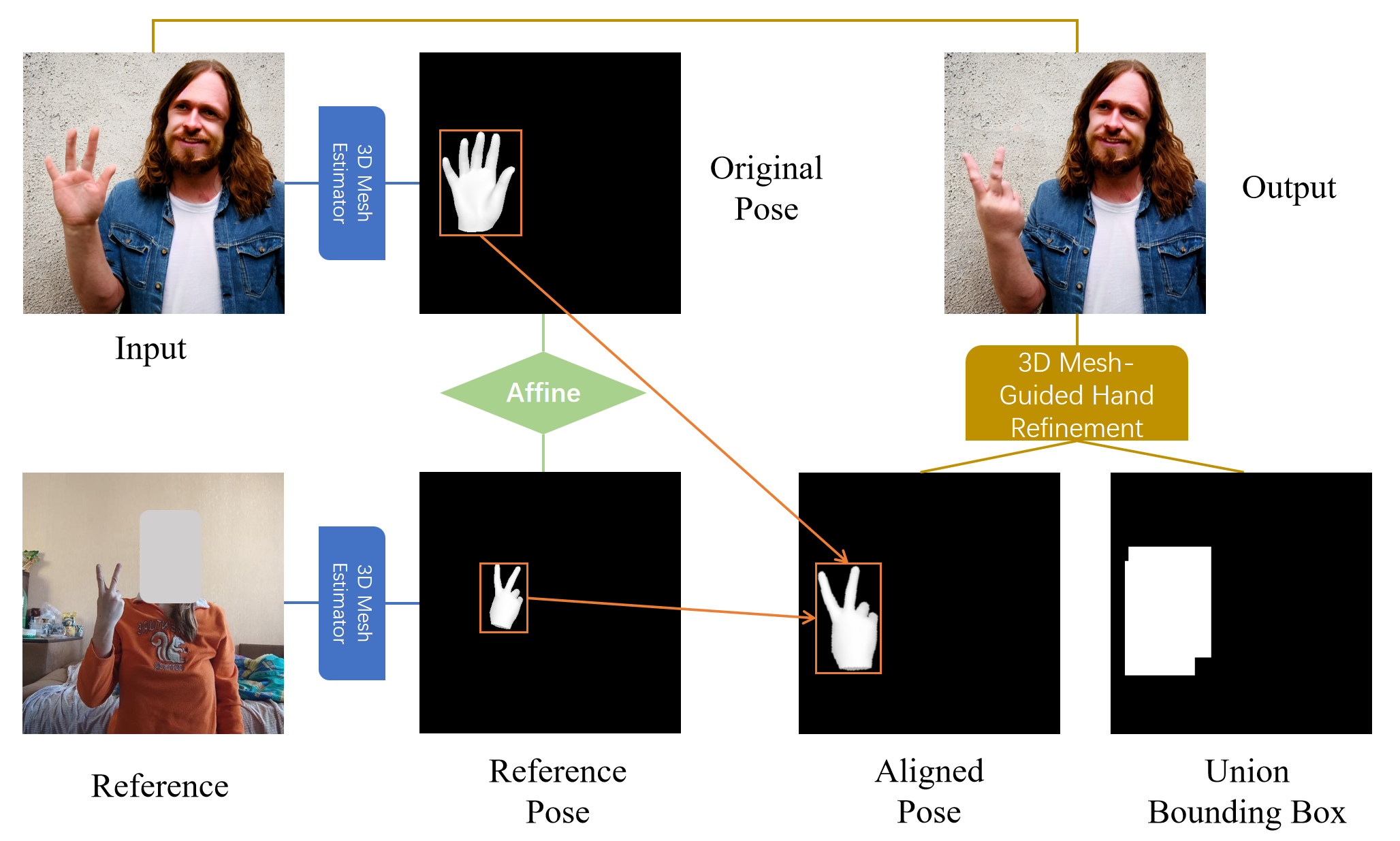}

  \caption{\textbf{The illustration of the hand pose transformation process.} We use a 3D hand mesh estimator to extract hand pose, then use an affine transformation to align pose location and scale, and then refine hands with our refinement method.}
  \label{fig:trans} 
    \vspace{10mm}
\end{figure*}

\subsection{Hand Pose Transformation}
The hand pose transformation process is illustrated in Figure \ref{fig:trans}. The framework is in inference mode. We have an input image containing a malformed hand and a reference image for our target pose. Firstly, we use Interwild \cite{moon2023bringing} to predict the 3D hand meshes for both the input image and the reference image. Now, we have the original pose and the reference pose. To transform the pose, we need to align the two hand poses. It involves three aspects: translation of the position, change in hand scale, and change in wrist angle. Motivated by Handcraft \cite{qin2024handcraft}, we use an affine transformation to do the gesture transformation. To calculate our alignment transformation matrix, we need two hand keypoints to help us. The keypoints we use are detected by Mediapipe \cite{zhang2020mediapipe}. We choose the wrist keypoint and the pinky finger keypoint. In practice, the wrist keypoint is unique for conducting rotation around the wrist. The other keypoint can be replaced by another keypoint besides the pinky finger. In detail, we illustrate the computation process.

\textbf{Scale} To compute the scale of two hand poses, we compute the distance between keypoint 1 and keypoint 2 in both hands. Then compute the ratio between the distance of pose 1 and pose 2. The result is our scale $s$. Using an affine transformation,  points in initial pose $\mathbf{x}=(x,y,1)^T$ can be represented as follows,

\begin{equation}
\mathbf{x}' = 
\begin{pmatrix}
s & 0 & 0 \\
0 & s & 0 \\
0 & 0 & 1
\end{pmatrix}
\begin{pmatrix}
x \\
y \\
1
\end{pmatrix} = 
\begin{pmatrix}
sx \\
sy \\
1
\end{pmatrix}
\end{equation}

where $\mathbf{x}'=(x',y',1)^T$ is the transformed point. The $2\times2$ submatrix $\begin{pmatrix}s&0\\0&s\end{pmatrix}$ controls the scale. 

\textbf{Translation} To compute the translation matrix that controls location movement. We compute the coordinate shift between keypoint 1 in the two hand poses. The shift is $(t_x,t_y)$. Points in pose after scale transformation $\mathbf{x}'=(x',y',1)^T$ can be represented as follows, 

\begin{equation}
\mathbf{x}''=\begin{pmatrix}
1&0&t_x\\
0&1&t_y\\
0&0&1
\end{pmatrix}
\begin{pmatrix}
x'\\
y'\\
1
\end{pmatrix}=
\begin{pmatrix}
x' + t_x\\
y' + t_y\\
1
\end{pmatrix}
\end{equation}

where $\mathbf{x}''=(x'',y'',1)^T$ is the transformed point. The $(t_x,t_y)$ controls the translation. 

\textbf{Rotation} To compute the rotation, we first translate the rotation center of the initial pose from keypoint 1 (wrist) to $(0,0)$.
Points in pose after scale and translation $\mathbf{x}''=(x'',y'',1)^T$ can be represented as follows, 

\begin{equation}
\mathbf{x}'''=\begin{pmatrix}
1&0&-c_x\\
0&1&-c_y\\
0&0&1
\end{pmatrix}
\begin{pmatrix}
x''\\
y''\\
1
\end{pmatrix}=
\begin{pmatrix}
x'' - c_x\\
y'' - c_y\\
1
\end{pmatrix}
\end{equation}

where $\mathbf{x}'''=(x''',y''',1)^T$ is the transformed point. $(c_x,c_y)$ is the coordinate of the keypoint 1. 

Then we compute the vector of keypoint 1 and keypoint 2 within two pose. And then, we compute the rotation $\theta$ of the two vectors. Points  $\mathbf{x}'''=(x''',y'',1)^T$ can be represented as follows, 

\begin{equation}
\mathbf{x}^4=\begin{pmatrix}
cos\theta&-sin\theta&0\\
sin\theta&cos\theta&0\\
0&0&1
\end{pmatrix}
\begin{pmatrix}
x'''\\
y'''\\
1
\end{pmatrix}=
\begin{pmatrix}
x'''cos\theta - y'''sin\theta\\
x'''sin\theta + y'''cos\theta\\
1
\end{pmatrix}
\end{equation}

where $\mathbf{x}^4=(x^4,y^4,1)^T$ is the transformed point. The $2\times2$ submatrix controls the rotation. Finally, we translate the center to keypoint 1 from $(0,0)$ as follows,

\begin{equation}
\mathbf{x}^5=\begin{pmatrix}
1&0&c_x\\
0&1&c_y\\
0&0&1
\end{pmatrix}
\begin{pmatrix}
x^4\\
y^4\\
1
\end{pmatrix}=
\begin{pmatrix}
x^4 + c_x\\
y^4 + c_y\\
1
\end{pmatrix}
\end{equation}

where $\mathbf{x}^5=(x^5,y^5,1)^T$ is the final coordinates of the transformed point. We use this translation to transform the target hand pose to the initial pose. Then, we use our proposed 3D hand mesh-guided hand refinement method to generate the final output result. It requires an aligned pose, a bounding box, and an input image as inputs. Note that the bounding box needs to be the maximum bounding box, which is the union region of the original gesture and the aligned gesture. It can guarantee that the original hand pose region and the new hand pose region are both covered.

\section{Experiments}\label{sec:experiments}

\subsection{Datasets}
\begin{figure}[h]
    \centering
    \includegraphics[width=12cm]{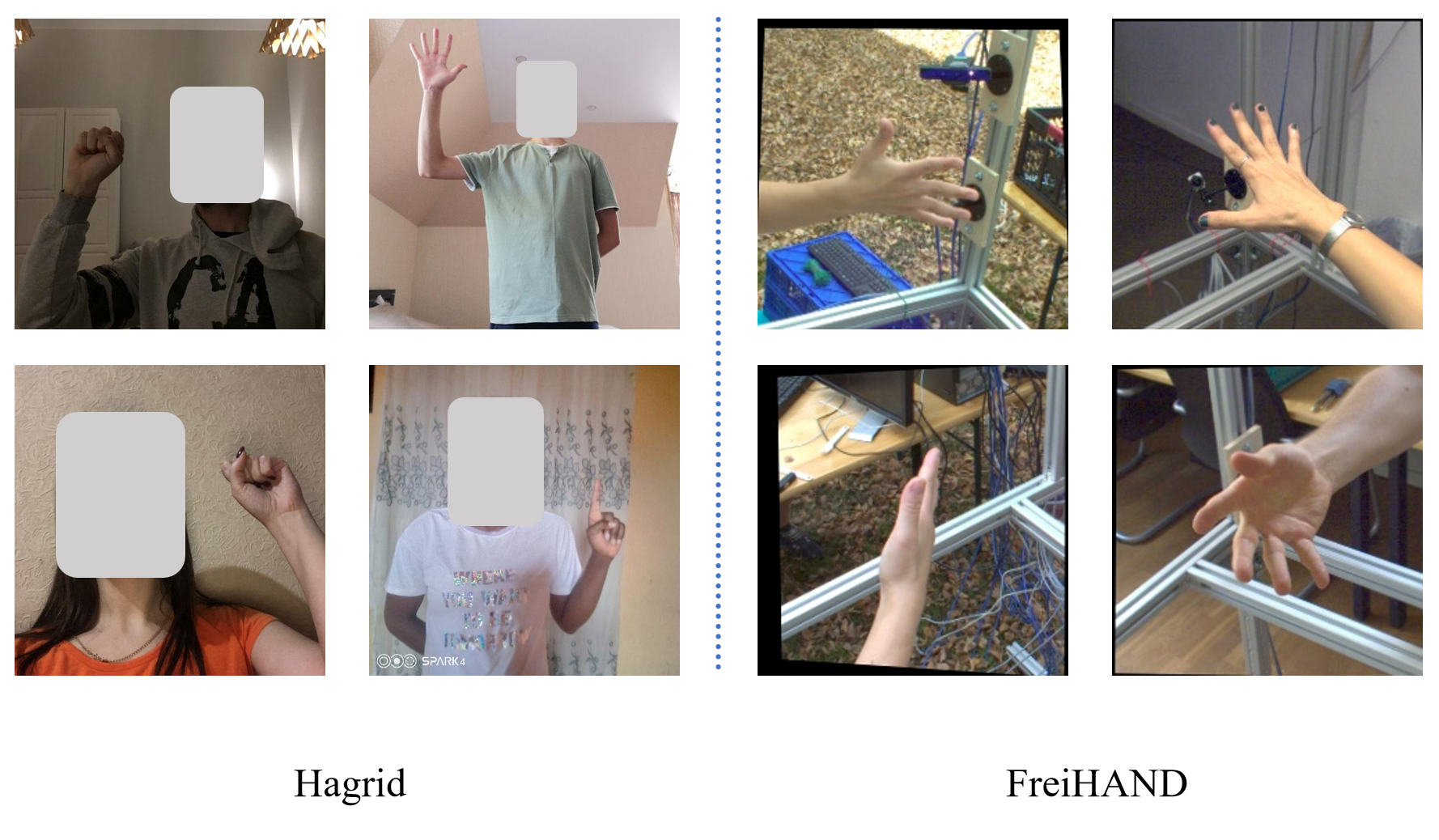}
    \caption{\textbf{Visualization of the reference dataset: Hagrid~\cite{hagrid} and FreiHAND~\cite{freihand}.} The faces of the real persons in the pictures are obscured due to privacy protection.}
    \label{fig:hagridandfrei}
    \vspace{10mm}
\end{figure}

We use our proposed training dataset for model training. It contains 24,411 triplets, each of which includes hand images, hand meshes, and hand region masks. For the evaluation datasets, we adopt the Hagrid~\cite{hagrid} and FreiHAND~\cite{freihand} datasets following HandRefiner \cite{lu2024handrefiner}. They are used to evaluate the generation quality of different hand-refinement models. A visualization of the two datasets is shown in Figure \ref{fig:hagridandfrei}. Hagrid is a large dataset including 0.55M hand and human body RGB images. It contains 18 common hand gestures. Every image includes a human with different hand gestures. FreiHAND consists of 32,560 samples with different hand poses. 

\subsection{Evaluation metrics} 
Following the baseline model HandRefiner \cite{lu2024handrefiner}, we adopt the Frechet Inception Distance (FID)~\cite{heusel2018gans} and the Kernel Inception Distance (KID)~\cite{bińkowski2021demystifying} on refined results and reference datasets, FreiHAND and Hagrid. We also use the keypoint detection confidence scores of a hand detector~\cite{zhang2020mediapipe} to evaluate the realness of the hands generated.

\subsection{Implementation details}
We implement our model using the Stable Diffusion model \cite{rombach2022high}. We froze the backbone parameters and finetune the ControlNet \cite{zhang2023adding} branch. The ControlNet branch is initialized with the depth-guided image generation weights. Our fine-tuned training steps are 2,307. We use the AdamW optimizer~\cite{adamW} with a batch size of 4. Our learning rate is 2e-5. The training and testing of models are conducted on a single NVIDIA A40 GPU. We use Interwild \cite{moon2023bringing} for hand mesh generation during inference. 

\subsection{Performance}
We compare our model with the original generation results of Stable Diffusion \cite{rombach2022high} and the existing malformed hand refinement methods. The methods include depth-guided ControlNet \cite{zhang2023adding} with depth maps generated by Graphormer \cite{lin2021mesh}, and HandRefiner with depth maps generated by Graphormer \cite{lin2021mesh}. Our method uses 3D hand mesh generated by Interwild \cite{moon2023bringing}. 

\begin{figure}[h]
    \centering
    \includegraphics[width=\columnwidth]{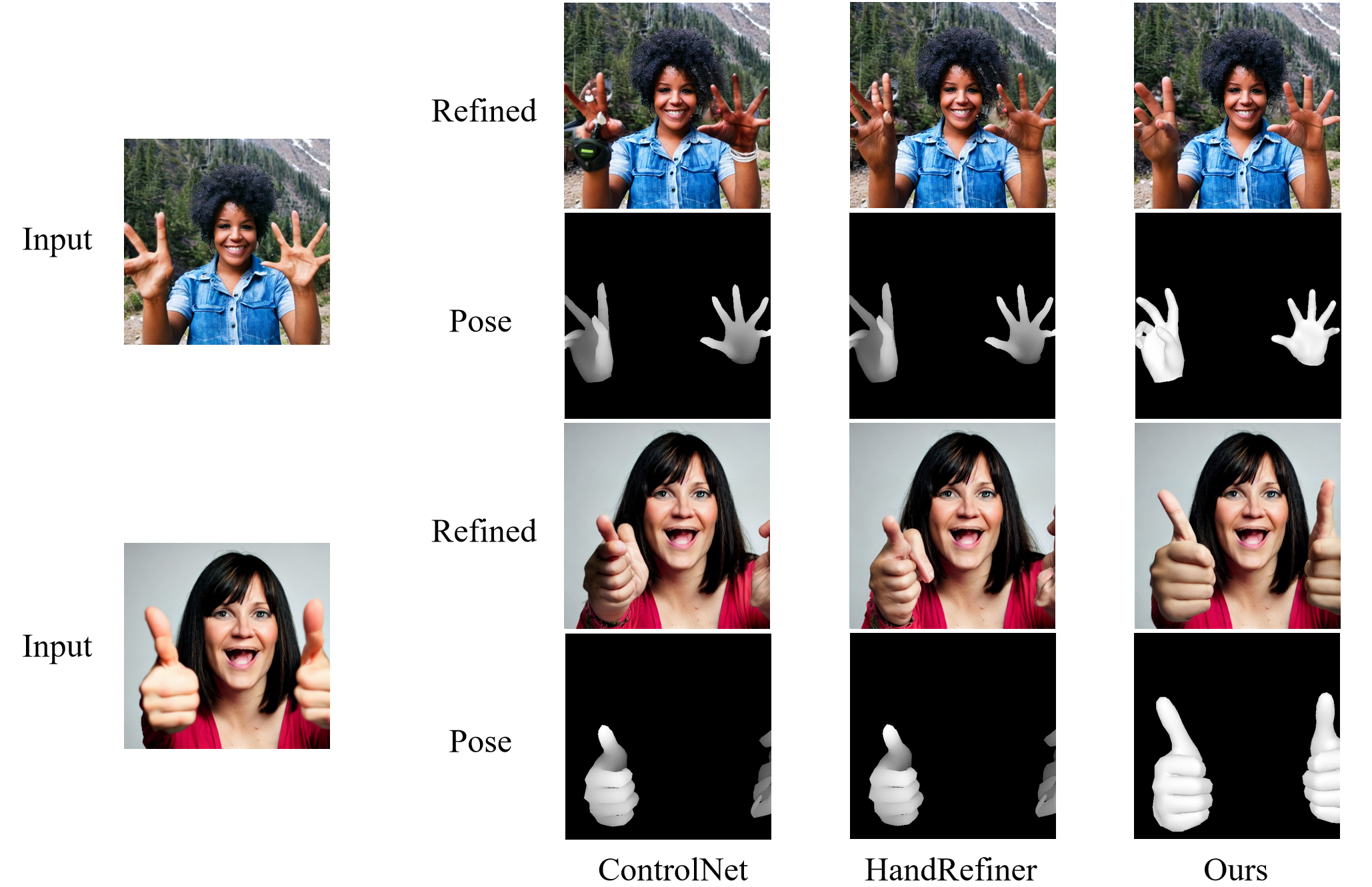}
    \caption{\textbf{Visual results on the Hagrid dataset.} The results have the whole human body with hands.}
    \label{fig:hagridvisual}
    \vspace{10mm}
\end{figure}

\begin{figure}[h]
    \centering
    \includegraphics[width=\columnwidth]{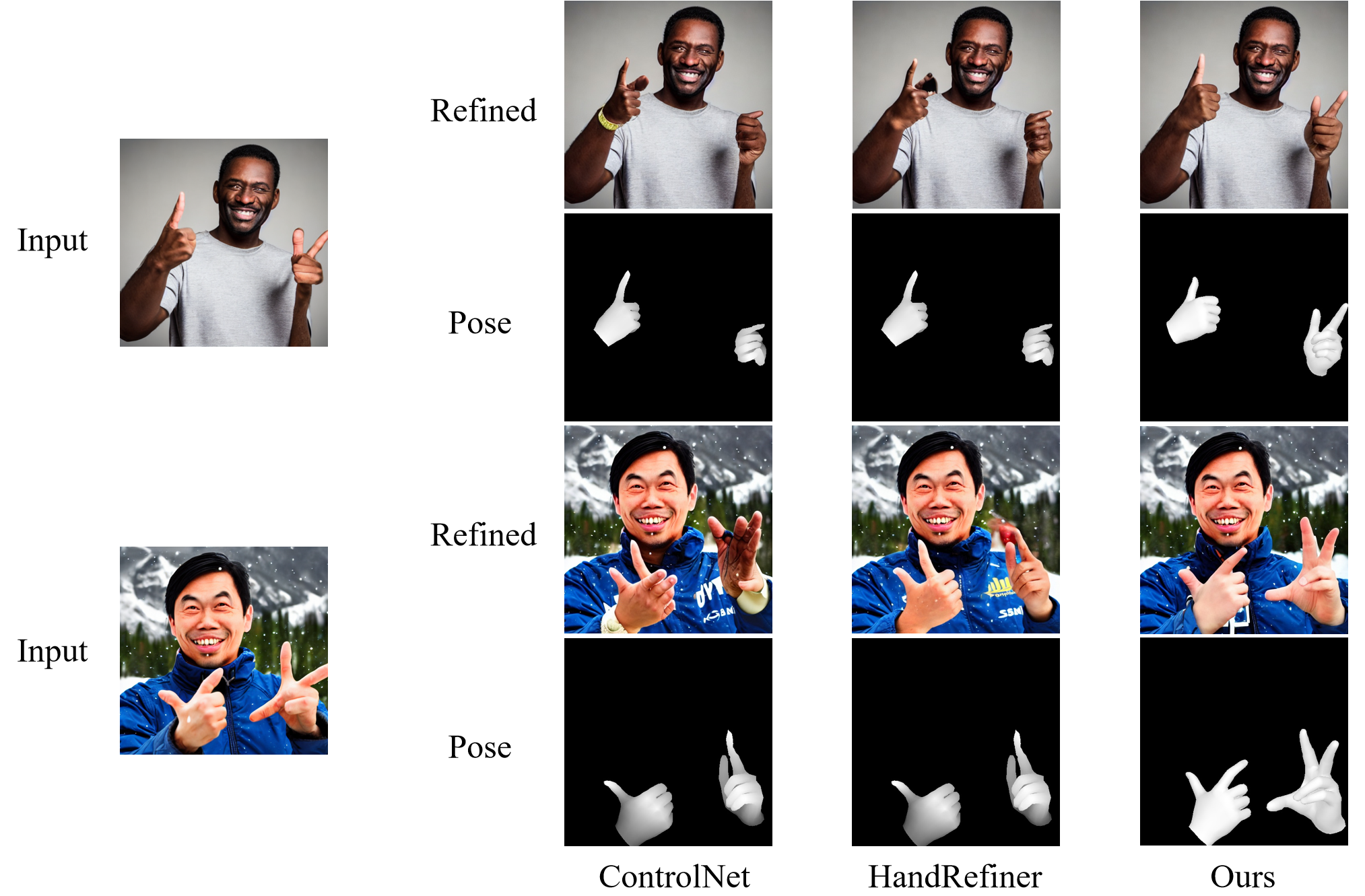}
    \caption{\textbf{Visual results on the Hagrid dataset.} The results have the whole human body with hands.}
    \label{fig:hagridvisual2}
    \vspace{10mm}
\end{figure}

\noindent\textbf{Hagrid as the Reference.}
Following HandRefiner \cite{lu2024handrefiner}, we generate 12K images for evaluation using the text descriptions comes from the Hagrid dataset. The text description covering 18 diverse gestures. We compose 12K unique text descriptions, such as 'an asian woman with ponytail facing the camera, making a gesture of calling, indoor'. Ethnicity, gender, background, hairstyle, and hand gesture can be substituted in the text descriptions. We then input the descriptions into the Stable Diffusion model and generate images. We process these images with ControlNet, HandRefiner, and our method separately. The comparison results are shown in Table~\ref{tab:hagrid}. It can be seen in the table that the results of our method have a lower FID/KID score compared to the original stable diffusion, results by ControlNet, and results by HandRefiner. For detection confidence, our method also outperforms other methods. Visual comparisons are shown in Figure \ref{fig:hagridvisual} and Figure \ref{fig:hagridvisual2}. Our 3D hand mesh maps exhibit significantly finer details compared to hand depth maps. At the same time, our generated refinement results are better than previous methods.

\begin{table}
  \centering
  \begin{tabular}{l c c c}
  \hline
    Method & FID $\downarrow$ & KID $\downarrow$ & Det. Conf. $\uparrow$ \\
          \hline 
    Stable Diffusion \cite{rombach2022high} & 81.58 & 0.077 & 0.90  \\
    ControlNet \cite{zhang2023adding} & 80.87 & 0.075 & 0.91 \\
    HandRefiner \cite{lu2024handrefiner} & 80.67 & 0.075 & 0.91  \\
    \textbf{Ours} & \textbf{80.11} & \textbf{0.074} & \textbf{0.92} 
 \\
      \hline 
  \end{tabular}
  \caption{Comparison of FID, KID, and detection confidence between original images by Stable Diffusion and rectified images by previous methods and our method using Hagrid as the reference.}
  \label{tab:hagrid}
      \vspace{20pt}
\end{table}



\begin{figure}
    \centering
    \includegraphics[width=11cm]{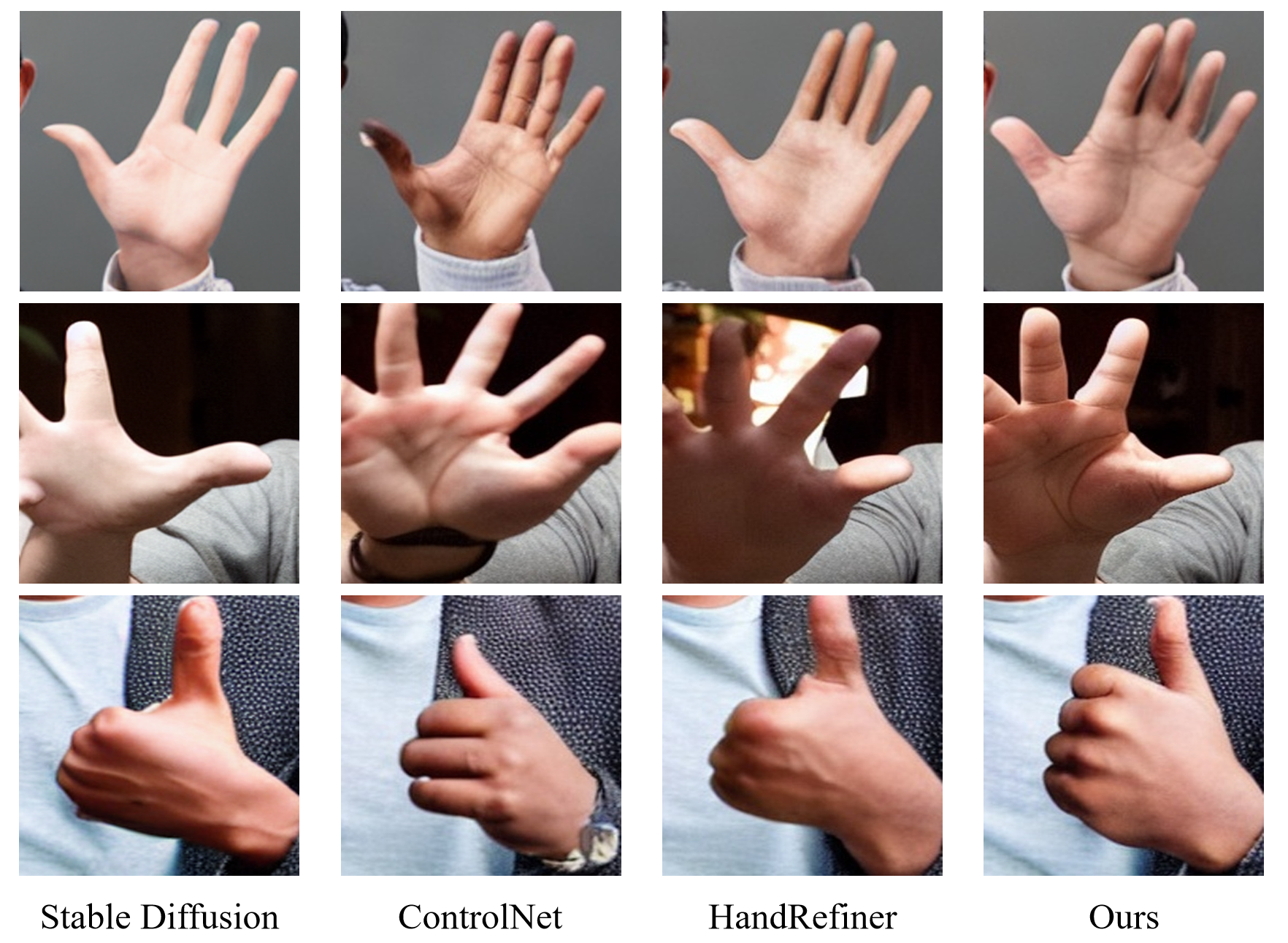}
    \caption{\textbf{Visual results on the FreiHAND dataset.} The results are cropped hand regions.}
    \label{fig:visualfreihand}
    \vspace{10mm}
\end{figure}

\noindent\textbf{FreiHAND as the Reference.}
The FreiHand dataset contains pure hands instead of the human body and hands. Following HandRefiner \cite{lu2024handrefiner}, we crop the 12K images generated by the stable diffusion model using the hand region mask by Interhand \cite{moon2023bringing}, then resize the hand images to 512 $\times$ 512. Then we used the cropped images for evaluation. We separately obtained 12,442 cropped hand images for the original results of Stable Diffusion, the results of ControlNet, the results of HandRefiner, and the results of our method. The crop masks are the same. For the results shown in Table~\ref{tab:freihand}, we can see that our method outperforms both stable diffusion and other refinement methods, in terms of FID and KID. Visual comparisons are shown in Figure \ref{fig:visualfreihand}. We can see that our generated refinement results are better than previous methods.

\begin{table}
\centering
  \begin{tabular}{l c c}
    \hline
    Method & FID $\downarrow$ & KID $\downarrow$ \\
      \hline 
    Stable Diffusion \cite{rombach2022high} & 130.12 & 0.090 \\
    ControlNet \cite{zhang2023adding} & 127.13 & 0.086 \\
    HandRefiner \cite{lu2024handrefiner} & 126.83 & 0.085 \\
    \textbf{Ours} & \textbf{125.75} & \textbf{0.084}   \\
      \hline
  \end{tabular}
  \caption{Comparison of FID and KID between original images, refined images by previous refinement methods, and our methods using FreiHAND as the reference.}
  \label{tab:freihand}
  \vspace{10pt}
\end{table}

\begin{figure}[h]
    \centering
    \includegraphics[width=\columnwidth] {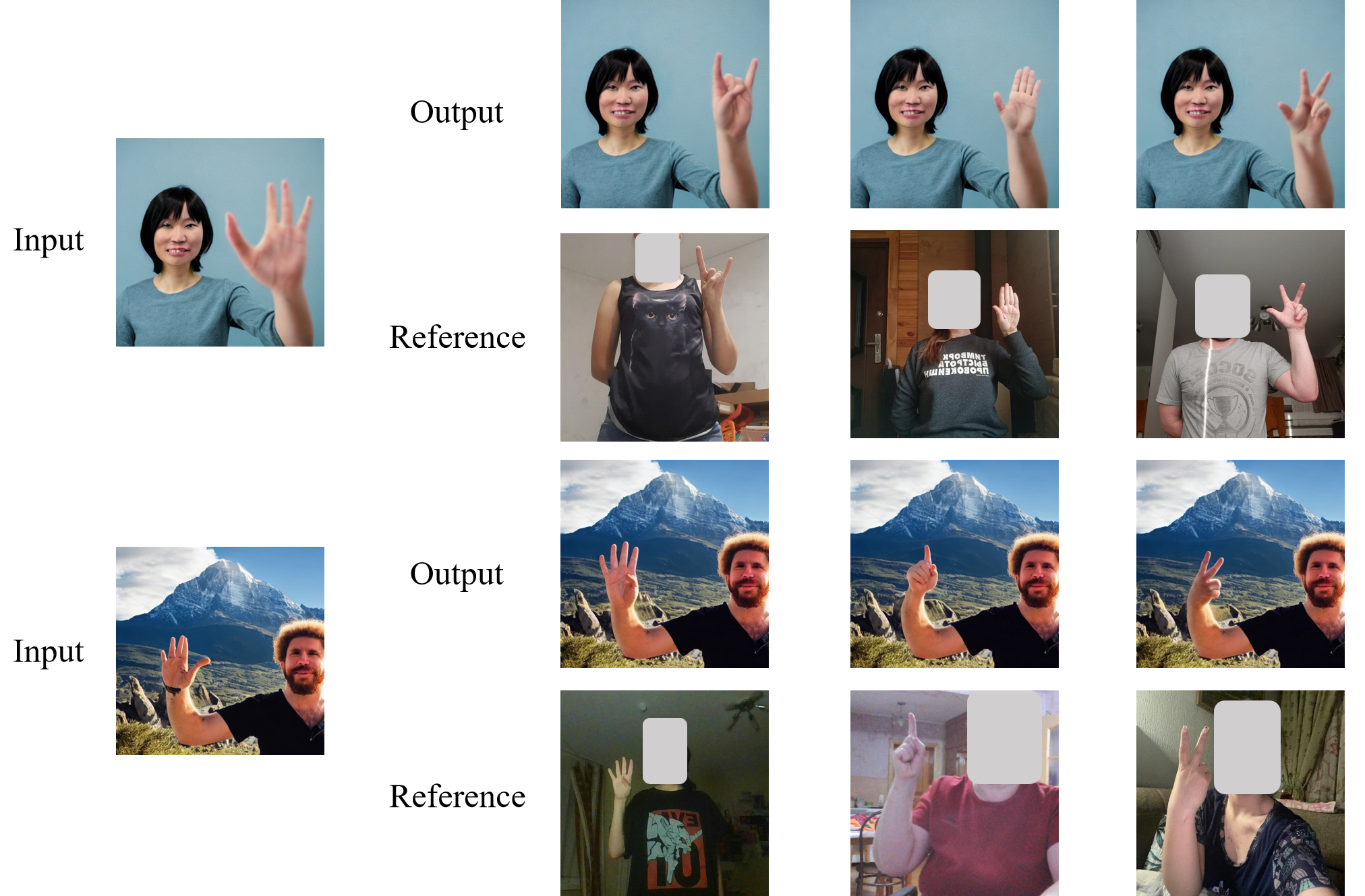}
    \caption{\textbf{Visual results of pose transformation.} The faces are masked for privacy protection.}
    \label{fig:posevisual}
    \vspace{10mm}
\end{figure}

\noindent\textbf{Pose Transformation.}
We provide a pose transformation experiment using human pose images from the Hagrid dataset. The results are shown in Figure \ref{fig:posevisual}. We can see that our pose transformation method can imitate the hand pose of the reference images.

\subsection{Ablation Studies}
\subsubsection{Double Check}
\begin{figure}
    \centering
    \includegraphics[width=12cm]{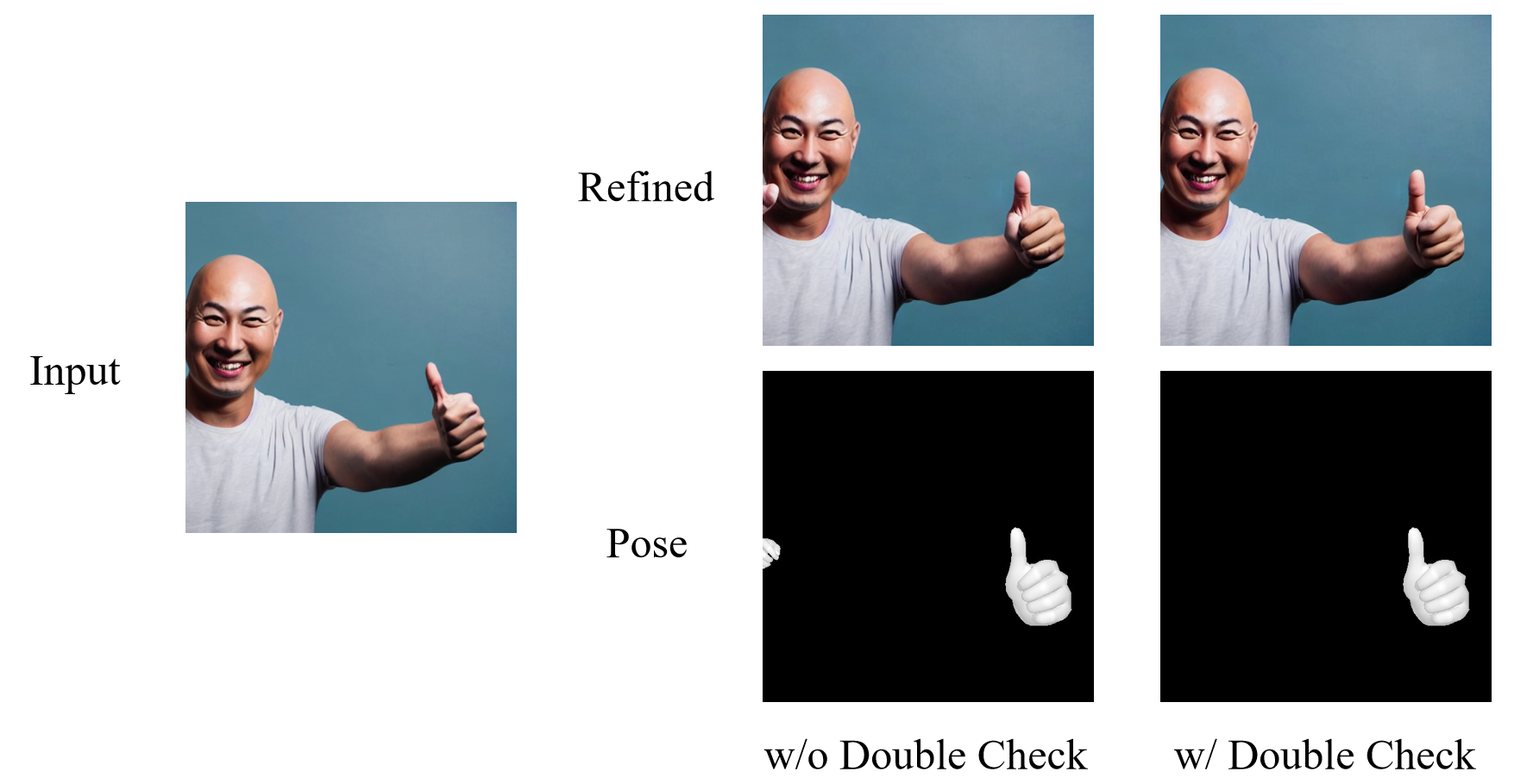}
    \caption{\textbf{Visual results from the ablation study of the double-check mechanism.} It clearly demonstrates the mechanism’s ability to eliminate wrong hand predictions.}
    \label{fig:abladouble}
    \vspace{10mm}
\end{figure}

We evaluate the performance of our method without using the double-check algorithm for hand mesh pre-processing during inference. The results are shown in Table \ref{tab:abladouble}. The refinement results generated by our method with the double check algorithm exhibit notable improvements in FID, KID, and detection confidence metrics, underscoring the algorithm's efficacy in enhancing output quality. A visual result is shown in Figure \ref{fig:abladouble}.

\begin{table}
\centering
  \begin{tabular}{l c c c}
    \hline
    Method & FID $\downarrow$ & KID $\downarrow$ & Det. Conf.$\uparrow$\\
      \hline
    w/ Double Check & 80.41 & 0.075 & 0.91\\
    \textbf{Ours} & \textbf{80.11} & \textbf{0.074}  & \textbf{0.92} \\
      \hline
  \end{tabular}
  \caption{Ablation study of double check algorithm.}
  \label{tab:abladouble}
    \vspace{10mm}
\end{table}

\subsubsection{Sub-dataset}
We evaluate the performance of our method using different combinations of our three sub-dataset. First, we use the InterHand dataset only for training. Then we combine InterHand and ReIH for training. Finally, we use InterHand, ReIH and Fashion for training. As evidenced by the results in Table \ref{tab:datasetabla}, the generative performance improves with the inclusion of more sub-datasets during training. This trend underscores the importance of our three sub-datasets in enhancing the refinement quality.

\begin{table}
\centering
  \begin{tabular}{l c c c}
    \hline
    Training Dataset & FID $\downarrow$ & KID $\downarrow$  & Det. Conf. $\uparrow$\\
      \hline
    InterHand & 80.55 & 0.075 & 0.91\\
    InterHand + ReIH & 80.37 & 0.074 & \textbf{0.92}\\
    InterHand + ReIH + Fashion (Ours)& \textbf{80.11} & \textbf{0.074} & \textbf{0.92}\\
      \hline
  \end{tabular}
  \caption{\textbf{Ablation study of different sub-datasets involved in training.}}
  \label{tab:datasetabla}
  \vspace{10pt}
\end{table}

\section{Conclusion and Future Work}
In this paper, we proposed a 3d mesh-guided diffusion-based malformed hand refinement pipeline for AI-generated images. This method outperforms the previous hand depth-guided refinement method due to the superior hand structure estimation performance of the hand mesh estimator. It provides better details, such as the back of the hand and the palm of the hand. To support our method, we proposed a double check algorithm that can exclude some wrong hand mesh predictions. It makes the model more robust. In addition, we proposed a hand pose transformation algorithm to help users change the generated hand pose. It makes the refinement more flexible. Extensive experiments have verified the superiority of our method. Meanwhile, our method currently has limitations in generating more complex scenes, such as occluded hands and hand-object interactions. Addressing these challenges will be a key focus of our future work.

\bibliographystyle{elsarticle-num}
\bibliography{reference}

\begin{thebibliography}{10}
\expandafter\ifx\csname url\endcsname\relax
  \def\url#1{\texttt{#1}}\fi
\expandafter\ifx\csname urlprefix\endcsname\relax\def\urlprefix{URL }\fi
\expandafter\ifx\csname href\endcsname\relax
  \def\href#1#2{#2} \def\path#1{#1}\fi

\bibitem{rombach2022high}
R.~Rombach, A.~Blattmann, D.~Lorenz, P.~Esser, B.~Ommer, High-resolution image synthesis with latent diffusion models, in: Proceedings of the IEEE/CVF Conference on Computer Vision and Pattern Recognition, 2022, pp. 10684--10695.

\bibitem{lu2024handrefiner}
W.~Lu, Y.~Xu, J.~Zhang, C.~Wang, D.~Tao, Handrefiner: Refining malformed hands in generated images by diffusion-based conditional inpainting, in: Proceedings of the 32nd ACM International Conference on Multimedia, 2024, pp. 7085--7093.

\bibitem{lin2021mesh}
K.~Lin, L.~Wang, Z.~Liu, Mesh graphormer, in: Proceedings of the IEEE/CVF international conference on computer vision, 2021, pp. 12939--12948.

\bibitem{zhang2020mediapipe}
F.~Zhang, V.~Bazarevsky, A.~Vakunov, A.~Tkachenka, G.~Sung, C.~Chang, M.~Grundmann, Mediapipe hands: On-device real-time hand tracking, ArXiv Preprint ArXiv:2006.10214 (2020).

\bibitem{song2020denoising}
J.~Song, C.~Meng, S.~Ermon, Denoising diffusion implicit models, arXiv preprint arXiv:2010.02502 (2020).

\bibitem{zhang2023adding}
L.~Zhang, A.~Rao, M.~Agrawala, Adding conditional control to text-to-image diffusion models, in: Proceedings of the IEEE/CVF International Conference on Computer Vision, 2023, pp. 3836--3847.

\bibitem{zhou2024adaptive}
Y.~Zhou, J.~Qian, H.~Zhang, X.~Xu, H.~Sun, F.~Zeng, Y.~Zhou, Adaptive multi-text union for stable text-to-image synthesis learning, Pattern Recognition 152 (2024) 110438.

\bibitem{song2025attridiffuser}
W.~Song, Z.~Ye, M.~Sun, X.~Hou, S.~Li, A.~Hao, Attridiffuser: Adversarially enhanced diffusion model for text-to-facial attribute image synthesis, Pattern Recognition (2025) 111447.

\bibitem{zhen2025token}
T.~Zhen, J.~Cao, X.~Sun, J.~Pan, Z.~Ji, Y.~Pang, Token-aware and step-aware acceleration for stable diffusion, Pattern Recognition 164 (2025) 111479.

\bibitem{qin2024handcraft}
Z.~Qin, Y.~Zhang, Y.~Liu, D.~Campbell, Handcraft: Anatomically correct restoration of malformed hands in diffusion generated images, ArXiv Preprint ArXiv:2411.04332 (2024).

\bibitem{zhang2024hand1000}
H.~Zhang, B.~Zhu, Y.~Cao, Y.~Hao, Hand1000: Generating realistic hands from text with only 1,000 images, ArXiv Preprint ArXiv:2408.15461 (2024).

\bibitem{park2024attentionhand}
J.~Park, K.~Kong, S.~Kang, Attentionhand: Text-driven controllable hand image generation for 3d hand reconstruction in the wild, in: European Conference on Computer Vision, 2024, pp. 329--345.

\bibitem{kwon2024graspdiffusion}
P.~Kwon, H.~Joo, Graspdiffusion: Synthesizing realistic whole-body hand-object interaction, ArXiv Preprint ArXiv:2410.13911 (2024).

\bibitem{pang2024manivideo}
Y.~Pang, R.~Shao, J.~Zhang, H.~Tu, Y.~Liu, B.~Zhou, H.~Zhang, Y.~Liu, Manivideo: Generating hand-object manipulation video with dexterous and generalizable grasping, ArXiv Preprint ArXiv:2412.16212 (2024).

\bibitem{liang2024vton}
Y.~Liang, X.~Hu, B.~Jiang, D.~Luo, K.~Wu, W.~Han, T.~Jin, C.~Wang, Vton-handfit: Virtual try-on for arbitrary hand pose guided by hand priors embedding, ArXiv Preprint ArXiv:2408.12340 (2024).

\bibitem{wang2025rhands}
C.~Wang, P.~Liu, M.~Zhou, M.~Zeng, X.~Li, T.~Ge, B.~Zheng, Rhands: Refining malformed hands for generated images with decoupled structure and style guidance, in: Proceedings of the AAAI Conference on Artificial Intelligence, Vol.~39, 2025, pp. 7573--7581.

\bibitem{pelykh2024giving}
A.~Pelykh, O.~Sincan, R.~Bowden, Giving a hand to diffusion models: a two-stage approach to improving conditional human image generation, in: 2024 IEEE 18th International Conference on Automatic Face and Gesture Recognition (FG), 2024, pp. 1--10.

\bibitem{eum2025mghand}
T.~Eum, J.~Choi, T.~Kim, Mghand: Multi-modal guidance for authentic hand diffusion, ArXiv Preprint ArXiv:2503.08133 (2025).

\bibitem{wang2025realishuman}
B.~Wang, J.~Zhou, J.~Bai, Y.~Yang, W.~Chen, F.~Wang, Z.~Lei, Realishuman: A two-stage approach for refining malformed human parts in generated images, in: Proceedings of the AAAI Conference on Artificial Intelligence, Vol.~39, 2025, pp. 7509--7517.

\bibitem{fu2025handrawer}
Q.~Fu, X.~Chen, M.~Asad, S.~Yuan, C.~Oh, G.~Slabaugh, Handrawer: Leveraging spatial information to render realistic hands using a conditional diffusion model in single stage, ArXiv Preprint ArXiv:2503.02127 (2025).

\bibitem{gao2024progressively}
K.~Gao, X.~Liu, P.~Ren, H.~Chen, T.~Zhen, L.~Xie, Z.~Li, Y.~Yan, H.~Zhang, E.~Yin, Progressively global--local fusion with explicit guidance for accurate and robust 3d hand pose reconstruction, Knowledge-Based Systems 304 (2024) 112532.

\bibitem{romero2022embodied}
J.~Romero, D.~Tzionas, M.~J. Black, Embodied hands: Modeling and capturing hands and bodies together, arXiv preprint arXiv:2201.02610 (2022).

\bibitem{zimmermann2019freihand}
C.~Zimmermann, D.~Ceylan, J.~Yang, B.~Russell, M.~Argus, T.~Brox, Freihand: A dataset for markerless capture of hand pose and shape from single rgb images, in: Proceedings of the IEEE/CVF international conference on computer vision, 2019, pp. 813--822.

\bibitem{zimmermann2017learning}
C.~Zimmermann, T.~Brox, Learning to estimate 3d hand pose from single rgb images, in: Proceedings of the IEEE international conference on computer vision, 2017, pp. 4903--4911.

\bibitem{kapitanov2024hagrid}
A.~Kapitanov, K.~Kvanchiani, A.~Nagaev, R.~Kraynov, A.~Makhliarchuk, Hagrid--hand gesture recognition image dataset, in: Proceedings of the IEEE/CVF Winter Conference on Applications of Computer Vision, 2024, pp. 4572--4581.

\bibitem{synthesisai2023static}
S.~AI, Static gestures dataset, \url{https://synthesis.ai/static-gestures-dataset/}, data retrieved from Synthesis AI (2023).

\bibitem{tzionas2016capturing}
D.~Tzionas, L.~Ballan, A.~Srikantha, P.~Aponte, M.~Pollefeys, J.~Gall, Capturing hands in action using discriminative salient points and physics simulation, International Journal of Computer Vision 118 (2016) 172--193.

\bibitem{moon2020interhand2}
G.~Moon, S.-I. Yu, H.~Wen, T.~Shiratori, K.~M. Lee, Interhand2. 6m: A dataset and baseline for 3d interacting hand pose estimation from a single rgb image, in: Computer Vision--ECCV 2020: 16th European Conference, Glasgow, UK, August 23--28, 2020, Proceedings, Part XX 16, Springer, 2020, pp. 548--564.

\bibitem{moon2023dataset}
G.~Moon, S.~Saito, W.~Xu, R.~Joshi, J.~Buffalini, H.~Bellan, N.~Rosen, J.~Richardson, M.~Mize, P.~De~Bree, et~al., A dataset of relighted 3d interacting hands, Advances in Neural Information Processing Systems 36 (2023) 17689--17701.

\bibitem{moon2023bringing}
G.~Moon, Bringing inputs to shared domains for 3d interacting hands recovery in the wild, in: Proceedings of the IEEE/CVF conference on computer vision and pattern recognition, 2023, pp. 17028--17037.

\bibitem{wu2021fashion}
H.~Wu, Y.~Gao, X.~Guo, Z.~Al-Halah, S.~Rennie, K.~Grauman, R.~Feris, Fashion iq: A new dataset towards retrieving images by natural language feedback, in: Proceedings of the IEEE/CVF Conference on computer vision and pattern recognition, 2021, pp. 11307--11317.

\bibitem{lugmayr2022repaint}
A.~Lugmayr, M.~Danelljan, A.~Romero, F.~Yu, R.~Timofte, L.~Van~Gool, Repaint: Inpainting using denoising diffusion probabilistic models, in: Proceedings of the IEEE/CVF conference on computer vision and pattern recognition, 2022, pp. 11461--11471.

\bibitem{redmon2016you}
J.~Redmon, S.~Divvala, R.~Girshick, A.~Farhadi, You only look once: Unified, real-time object detection, in: Proceedings of the IEEE conference on computer vision and pattern recognition, 2016, pp. 779--788.

\bibitem{hagrid}
A.~Kapitanov, A.~Makhlyarchuk, K.~Kvanchiani, Hagrid - hand gesture recognition image dataset (2022).
\newblock \href {http://arxiv.org/abs/2206.08219} {\path{arXiv:2206.08219}}.

\bibitem{freihand}
C.~Zimmermann, D.~Ceylan, J.~Yang, B.~Russell, M.~Argus, T.~Brox, Freihand: A dataset for markerless capture of hand pose and shape from single rgb images, in: Proceedings of the IEEE International Conference on Computer Vision, 2019.

\bibitem{heusel2018gans}
M.~Heusel, H.~Ramsauer, T.~Unterthiner, B.~Nessler, S.~Hochreiter, Gans trained by a two time-scale update rule converge to a local nash equilibrium, in: Advances In Neural Information Processing Systems, 2017.

\bibitem{bińkowski2021demystifying}
M.~Bińkowski, D.~J. Sutherland, M.~Arbel, A.~Gretton, Demystifying mmd gans, in: International Conference on Learning Representations, 2018.

\bibitem{adamW}
I.~Loshchilov, F.~Hutter, Decoupled weight decay regularization, in: International Conference on Learning Representations, 2019.

\end{thebibliography}

\end{document}